%% file: root.tex
\documentclass[letterpaper, 10 pt, conference]{Settings/ieeeconf}  

\input{Settings/settings.tex}

\DeclareUnicodeCharacter{2009}{\,}

\begin{document}

\maketitle
\thispagestyle{fancy} 
\pagestyle{fancy} 

\begin{abstract}

Safety for physical human-robot interaction (pHRI) is a major concern for all application domains. While current standardization for industrial robot applications provide safety constraints that address the onset of pain in blunt impacts, these impact thresholds are difficult to use on edged or pointed impactors. The most severe injuries occur in constrained contact scenarios, where crushing is possible. Nevertheless, situations potentially resulting in constrained contact only occur in certain areas of a workspace and design or organisational approaches can be used to avoid them. What remains are risks to the human physical integrity caused by unconstrained accidental contacts, which are difficult to avoid while maintaining robot motion efficiency. Nevertheless, the probability and severity of injuries occurring with edged or pointed impacting objects in unconstrained collisions is hardly researched. In this paper, we propose an experimental setup and procedure using two pendulums modeling human hands and arms and robots to understand the injury potential of unconstrained collisions of human hands with edged objects. Pig feet are used as ex vivo surrogate samples - as these closely resemble the physiological characteristics of human hands - to create an initial injury database on the severity of injuries caused by unconstrained edged or pointed impacts. For the effective mass range of typical lightweight robots, the data obtained show low probabilities of injuries such as skin cuts or bone/tendon injuries in unconstrained collisions when the velocity is reduced to $<$ \SI{0.5}{m/s}. Additionally, distinct differences between injury probability of the finger substitutes and the back of the hand substitutes are observed. The proposed experimental setups and procedures should be complemented by sufficient human modeling, e.g. the effective masses of human body parts, and will eventually lead to a complete understanding of the biomechanical injury potential in pHRI.


\end{abstract}

\glsresetall 
\section{Introduction}
 
 \begin{figure}[th]
	
		\includegraphics[width=1\linewidth]{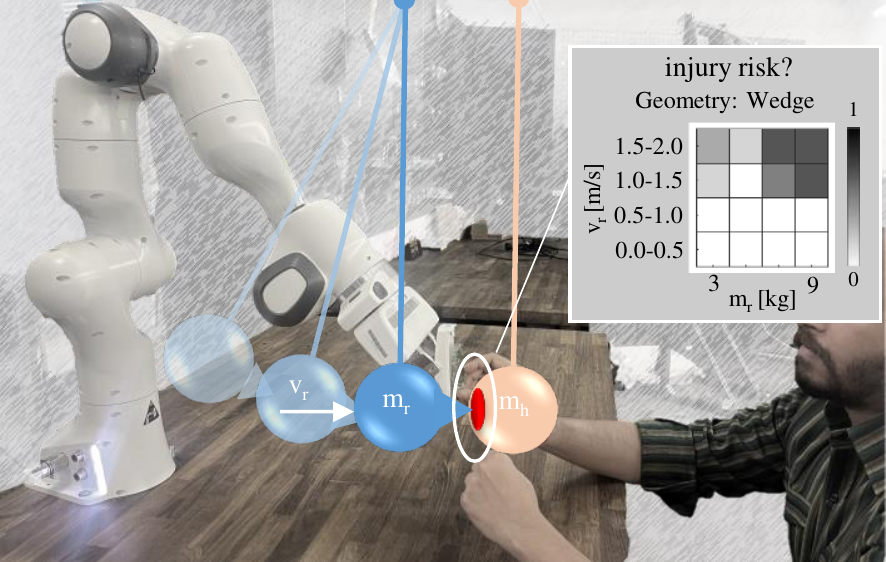}
	\caption{ To structurally investigate the severity and probability of occurrence of injuries occurring in unconstrained collision situations, we model the collision scenario using a defined impact geometry, the human effective body part mass, $m_h$, robot effective mass, $m_r$, and the robot's velocity, $v_r$.  }
	\label{fig:specimen}
	
	\vspace{-3mm}
	\label{fig:intro}
\end{figure}

Safety research in the field of physical human-robot interaction (pHRI) over the last century has been about limiting robot speeds or power and force based on the severity of injuries sustained in a collision \cite{Laffranchi_2009, iso15066, Haddadin.2012, Behrens_2014, Meguenani_2017}. To investigate injury occurrence, collision cases in which the human body part is constrained serve as the worst case as the body part cannot recoil and may even be clamped after the collision resulting in high injury probability and severity \cite{Haddadin.2009_requ}. Consequently, most existing injury studies that serve as the foundation of norms and regulations investigate contact between a robot and a human body part (or surrogate) that is restricted by a wall, for example in a way that it cannot strike back \cite{Han_2022,iso15066, Behrens_2014,Haddadin.2008_constr, Fujikawa_2013, Fujikawa_2023, Sugiura_2022}. Few researchers additionally conducted unconstrained collision tests \cite{Haddadin_2009, Haddadin.2008, Han_2021}.  

To reduce the probability of collisions between humans and robots, for example, constraints on the workspace of robots such as Cartesian regions and safely bounded positions (“virtual walls”) can be used \cite{Michalos_2015,Hjorth_2020}. While these are typically implemented in structured industrial environments as fixed constraints for rigid obstacle avoidance, research also focuses on creating more adaptive boundaries of this kind, based e.g. on the artificial potential fields approach \cite{Warren_1989, Janabi-Sharifi_1993, Vadakkepat_2000} or control barrier functions \cite{Ferraguti_2022}. These can be used to keep the robot in a safe area and thus prevent physical contact with humans, for example. Consequently, applications can be designed with a low probability of human-robot collisions, especially, in the constrained space, shifting the focus of risk assessment to scenarios with unconstrained collisions.
Using the resulting thresholds from constrained injury experiments as a worst-case assumption for even unconstrained contacts is an excellent basis for ensuring human safety when interacting with robots, but also limits robot efficiency by potentially making overly conservative assumptions about the injury severity that will occur \cite{Haddadin.2009_requ}. However, in order to enable efficient robot use and not to assess the risk in these collision scenarios too conservatively, structured studies on soft tissue injuries in unconstrained collision scenarios are required. 


In this paper, we present an experimental setup to structurally investigate the severity of damage caused under the assumption of collision under unconstrained conditions and conduct preliminary studies using pig feet as a substitute for human hands with different impactor geometries, effective masses of human body parts for collision with human arms, effective robot masses, and robot velocities.



\begin{table*}[t]
 
	\begin{minipage}[t]{\linewidth}
	
    \centering
    \caption{Overview of studies relevant to the field of robotics reporting severity of harm for collision scenarios a) or c)}
    \begin{tabular}{p{10mm}p{12mm}p{12mm}p{7mm}P{5mm}P{5mm}P{5mm}P{5mm}P{8mm}P{8mm}P{5mm}P{5mm}P{5mm}P{5mm}P{8mm}P{8mm}}
    \toprule
        \multirow{3}{*}{\textbf{Scenario}} & \multirow{3}{*}{\textbf{Injury}} & \multirow{3}{*}{\textbf{Bodypart}} & \multirow{3}{*}{\textbf{Studies}} & 
        \multicolumn{6}{c}{\textbf{Subjects}} & \multicolumn{4}{c}{\textbf{Injury parameters}}  & \multicolumn{2}{c}{\textbf{Impactor types}}  \\ 
        & & & & \multicolumn{2}{c}{human} & \multicolumn{2}{c}{surrogate} & \multirow{2}{*}{syn.} & \multirow{2}{*}{sim.} & \multirow{2}{*}{$F$} & \multirow{2}{*}{$p$} & \multirow{2}{*}{$[m,v]$} & \multirow{2}{*}{$E$}  & \multirow{2}{*}{edged} & \multirow{2}{*}{blunt}  \\
         & & & &  iv & ev & iv & ev & & \\
         
        \midrule

        \multirow{5}{*}{\textbf{a)}} & \parbox{12mm}{onset of pain} & \parbox{12mm}{head/leg/
        shoulder} & \cite{Han_2021} &
         \checkmark & \checkmark & - & -& - & - & \checkmark & \checkmark & (\checkmark) & - & - & \checkmark \\
         \\

         & \multirow{3}{*}{\parbox{12mm}{bone/ concussion}} & \multirow{2}{*}{\parbox{10mm}{head/neck/ chest}} &\cite{Haddadin_2009} & \checkmark & - & - & - & - & -  & \checkmark & \checkmark & \checkmark & \checkmark & - & \checkmark \\
        & &
           & \cite{Haddadin.2008} & \checkmark & - & - & -  &  \checkmark & \checkmark & \checkmark & - & - & - & - & \checkmark \\
         & & \multirow{1}{*}{\parbox{12mm}{fingers}} & \cite{Asad_2023} &
         - & - & - & - & - & \checkmark & \checkmark & \checkmark & (\checkmark) & \checkmark & - & \checkmark \\
        
         
         \midrule
         
         
         \multirow{11}{*}{\textbf{c)}} & \multirow{5}{*}{\parbox{12mm}{onset of pain}} & all & \cite{Behrens_2014} & 
         \checkmark & - & - & - & - & - & \checkmark & \checkmark & \checkmark & \checkmark & - & \checkmark \\
         
         & & all & \cite{Behrens_2022} & 
         \checkmark & - & - & - & - & - &\checkmark & \checkmark & - & - & - & \checkmark \\
         
         & & all & \cite{Yamada_1996} & 
         \checkmark & - & - & - & - & - &  \checkmark & - & - & - & - & \checkmark \\

         
         & & forearm & \cite{Povse_2010} &
         \checkmark & - & - & -& - & - & \checkmark & - & - & \checkmark & \checkmark & \checkmark \\
         
         
        \\
         & \multirow{4}{*}{\parbox{12mm}{skin contusion}} & hand/arm & \cite{Behrens2023} 
         & \checkmark & - & - & - & - & -&  \checkmark & \checkmark & - & \checkmark & - & \checkmark \\
         & & finger & \cite{Fujikawa_2023} & - & - & \checkmark & - & - & - & \checkmark & - & - & - & - & \checkmark \\
         & & soft tissue & \cite{Sugiura_2022} & - & - & \checkmark & - & - & - &  \checkmark & \checkmark & \checkmark & \checkmark & - & \checkmark \\
         & & legs & \cite{Desmoulin_2011} & \checkmark & - & - &- &- & - & \checkmark & \checkmark & (\checkmark) & \checkmark & - & \checkmark \\ 
         \\

         & \multirow{1}{*}{\parbox{12mm}{bone /concussion}} & \multirow{1}{*}{\parbox{12mm}{head/neck/ chest}} & \cite{Fujikawa_2013} & - & - & - & - & \checkmark & - & \checkmark & - & \checkmark & - & - & \checkmark \\
         \\
         
         \bottomrule
    \end{tabular}
    \vspace{-2mm}
    \label{tab:literature}
    \end{minipage}
    \vspace{-0mm}
\end{table*}

This paper is structured as follows. Sec. \ref{sec:State of the Art} summarizes state of the art available datasets on human injury occurrence applicable for unconstrained robot collisions. Sec. \ref{sec:methodology} motivates and describes the conducted injury experiments. The results are presented and discussed in Sec. \ref{sec:results}. Sec. \ref{sec:Discussion} then outlines how these results can be applied and discusses the limitations and further required work on the presented studies. Lastly, Sec. \ref{sec:Conclusion} concludes the paper.

\section{Available Human Unconstrained Contact Injury Data}\label{sec:State of the Art}

\begin{figure}[t]
\centering
\includegraphics[width=1\linewidth]{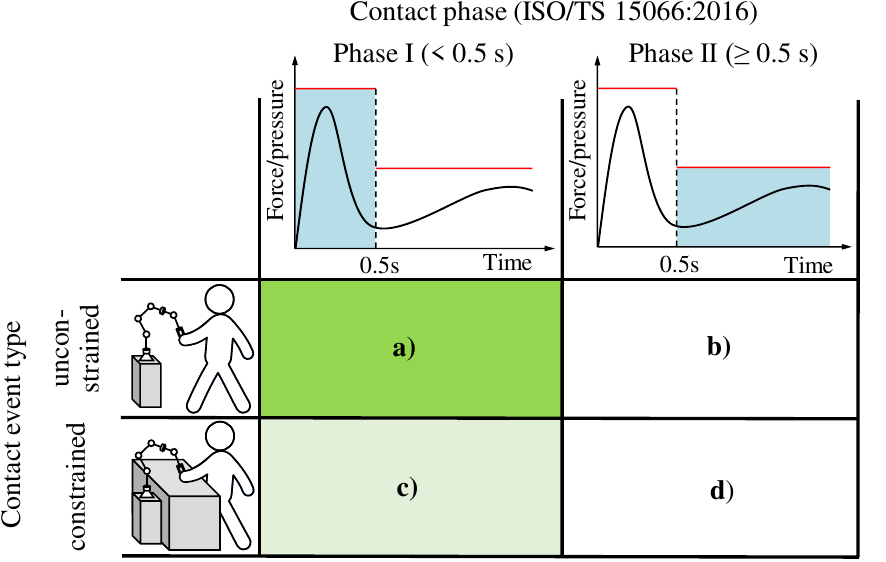}
\caption{Four collision scenarios: a) unconstrained and dynamic, b) unconstrained and quasi-static, c) constrained and dynamic, and d) constrained and quasi-static.}
\vspace{-0.3cm}
\label{fig:coll_scen}
\end{figure}

In pHRI, different collision situations may occur \cite{Haddadin.2009_requ}. To categorize them, we use the type of contact event, which describes the spatial constraints grouped into \emph{unconstrained} or \emph{constrained}\footnote{for the sake of simplicity semi-constrained and secondary impacts are excluded from the graphic}, and the temporal contact phase, which describes the contact duration, as depicted in Fig. \ref{fig:coll_scen}. The temporal distinction is based on ISO/TS 15066:2016 \cite{iso15066}. We refer to the first $<$ \SI{0.5}{s} of the impact as the first, \emph{dynamic} impact phase, and longer lasting contacts enter the second, \emph{quasi-static} phase once the duration is $\geq$ \SI{0.5}{s} \cite{iso15066}.

Researchers conducted various studies focusing on injury potential in these different collision situations \cite{Behrens_2014, Haddadin_2009, Haddadin.2008, Haddadin.2008_constr, Yamada_1996, Sugiura_2022}. In our study, we focus on the potential harm caused by collisions of type a); unconstrained in space and of short impact duration. As type c) collisions can be interpreted as the worst case to type a) collisions where the human effective mass becomes $m_h \approx \infty$ \cite{Svarny_2021}, we add available studies for these collision cases to the overview in Table \ref{tab:literature}. For each study, the overview table summarizes: a) the observed injuries, b) considered body parts, c) type of specimen including \emph{in vivo} (iv), \emph{ex vivo} (ev), and synthetic (syn.) specimen or simulations (sim.), d) which parameters were reported, such as force-based information, denoted as $F$, pressure $p$, mass, $m$, and velocity, $v$, or energy information, $E$, e) whether edged or blunt impactors were used. 
With this analysis of the state of the art injury protection datasets, the absence of data for edged impactor collisions becomes apparent. Moreover, only three studies address the actual unconstrained collision scenario, a). Two of these studies deal with severe head/neck or chest injuries, mainly using an anthropomorphic test device for car crash testing \cite{Haddadin.2008,Haddadin_2009}. The third study provides a dataset on human hand an finger bone strains obtained by simulation \cite{Asad_2023}. In the following, we therefore present a structured procedure to extend the injury protection data for scenario a) by experimental tests.



\section{Methodology} \label{sec:methodology}

 \begin{figure}[t]
\centering
\includegraphics[width=1\linewidth]{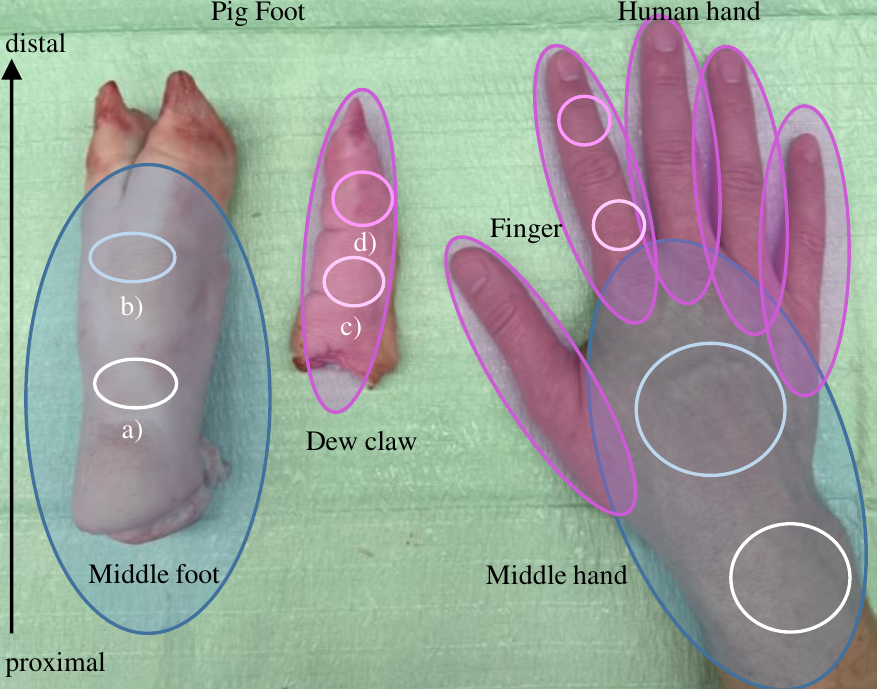}
\caption{Pig specimen for human hand substitution:  middle foot with a) proximal and b) distal collision location, representing the human middle hand and wrist region, dew claw with c) proximal and d) distal collision location, representing human fingers.}
\vspace{-0.3cm}
\label{fig:specimen}
\end{figure}

In this section, we describe the experimental focus, conditions, set up, and procedure of the structured unconstrained impact experiments. 

\subsection{Research Objectives}

The questions we would like to experimentally investigate are

\begin{itemize}
    \item[i)] which kind of injuries are observable in unconstrained light contact with edged or pointed impactors, 
    \item[ii)] how high is the probability of these injuries occurring,
    \item[iii)] what are the collision forces corresponding to the injury types, 
    \item[iv)] at which masses and velocities do the injuries occur, such that we can assemble mass-velocity based injury prevention datasets for robot control like in \cite{Haddadin.2012, Kirschner_2024_Towards, iso15066}, and
    \item[v)] how do the injury types differ between different effective arm masses also compared to previous constrained injury tests \cite{Kirschner_2024_Towards}.
\end{itemize}

As this is an exploratory experiment including a large number of different parameters, we focus on a descriptive analysis and do not aim for statistical tests. We base our choice of human surrogates on our previous work in \cite{Kirschner_2024_Towards}. There, we pointed out the anatomical similarity between pig paws and human hands and mentioned the difference in skin thickness considering dermis and epidermis of pig skin (between 1 and 6 mm \cite{Shergold2006}, \cite{Summerfield2015}) and the human skin on the back of the hand ($\sim$ \SI{2.3}{mm} \cite{Oltulu2018}).

Consequently, we use dissected pig dew claws and the back of pig feet as human finger and middle hand surrogates and investigate a distal and proximal collision location on each specimen as shown in Fig. \ref{fig:specimen}. 

\subsection{Impact Experiments}

In this study, we surrogate human hands and fingers using a pendulum and approximate the human effective body part mass based on ISO/TS 15066:2016 \cite{iso15066}.

\subsubsection{Experimental Setup Design}

\begin{table}[tpb]
    \centering
    \caption{Experimental settings}
    \vspace{-0.1cm}
    \begin{tabular}{p{11mm}p{13mm}p{27mm}p{18mm}}
        \toprule
        Parameter & & Value  & Based on   \\ 
        \midrule
        \multirow{2}{*}{\parbox{16mm}{$m$ [\SI{}{kg}]}} & impacted & $\sim$ \SI{2.6}{}, \SI{3.4}{} & hand/arm \cite{iso15066} \\
         &impacting & $\sim$ \SI{3}{}, \SI{5}{}, \SI{7}{}, \SI{9}{} &  robot reflected inertia \cite{Kirschner_2021_Notion}\\
         \parbox{16mm}{$v$ [\SI{}{m/s}]} &  & \SI{0.25}{}, \SI{0.5}{}, \SI{1.0}{}, \SI{1.5}{}, \SI{2.0}{} &  \cite{iso_10218-2}, \cite{franka_FR3_da}\\
         \multirow{3}{*}{\parbox{16mm}{impact geometries}} & wedge (W) & prism \ang{90}, boned &  \cite{Haddadin.2012}\\
          & edge (E)  & tetrahedron \ang{90}, boned &  \cite{Haddadin.2012}\\
           & sheet (S) & width \SI{1.5}{mm}, boned &  \cite{casalino_2018}\\
        \bottomrule
    \end{tabular}
    \label{tab:Experimental parameters}
    \vspace{-0.3cm}
\end{table}

\begin{figure}[t]
\centering
\includegraphics[width=1\linewidth]{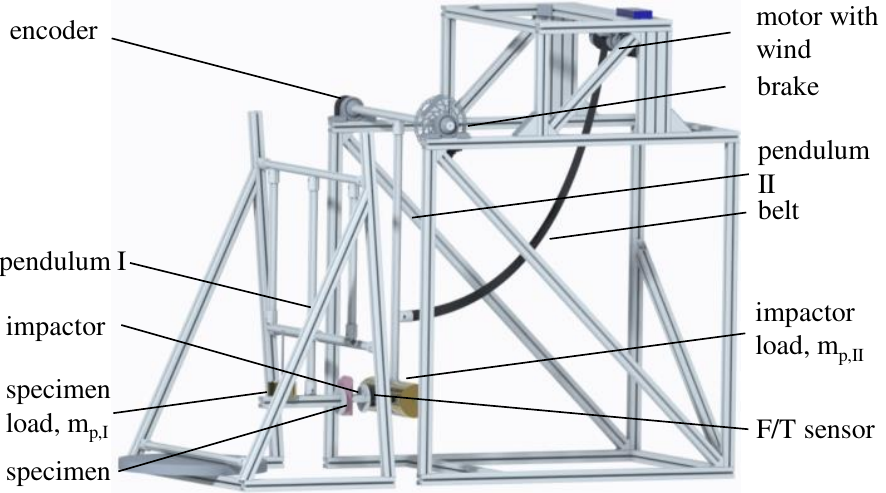}
\caption{Unconstrained impact test setup.}
\vspace{-0.3cm}
\label{fig:DT}
\end{figure}

\begin{figure*}[t]
\centering
\includegraphics[width=1\linewidth]{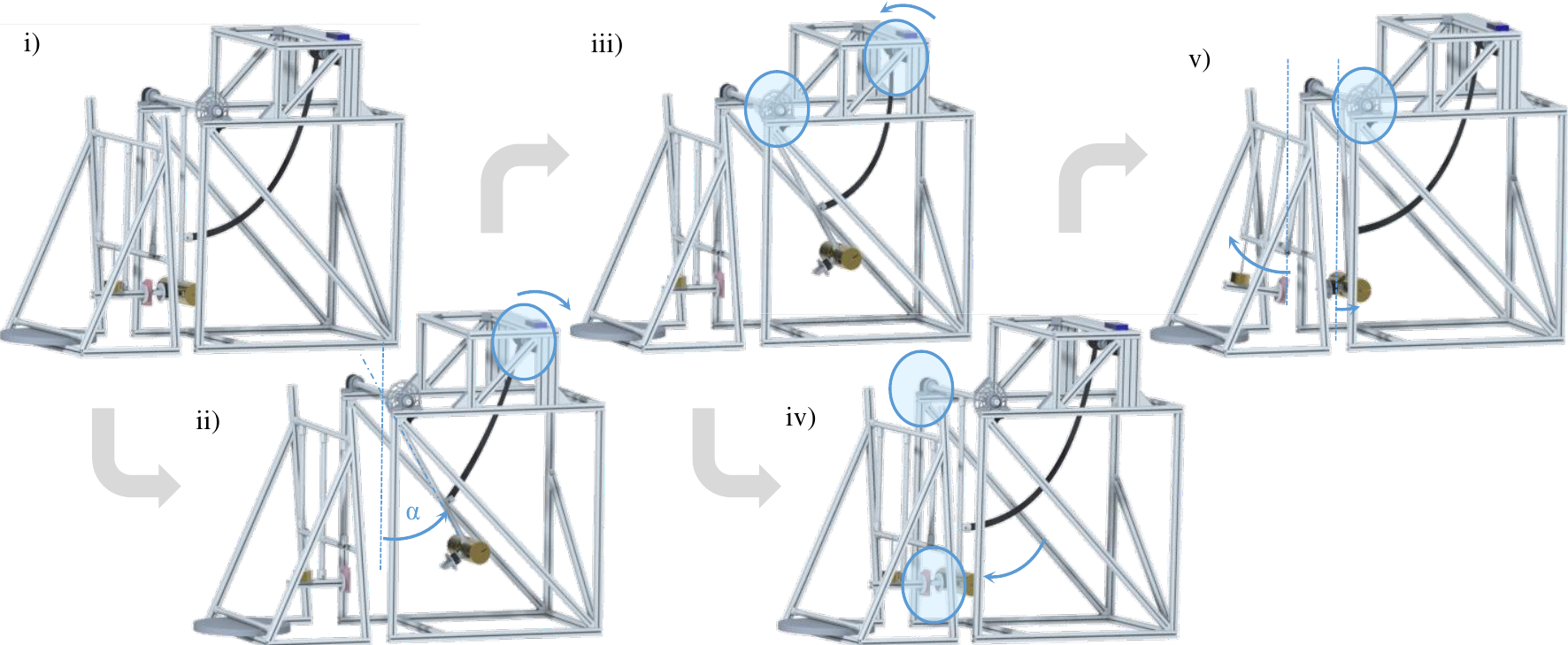}
\caption{Experimental procedure for unconstrained collision injury analysis. i) aligning specimen and impactor, ii) motor and wind drag pendulum-II to the desired angle $\alpha$, iii) pendulum is held by brake while motor unwinds the strap, iv) brake is released and pendulum-II collides with pendulum-I while encoder and force sensor measurement is active, v) pendulum-I swings and is caught by the human operator, pendulum-II is stopped by the brake when swinging back.}
\vspace{-0.2cm}
\label{fig:protocol_flowchart}
\vspace{-2mm}
\end{figure*}

The test rig is depicted in Fig. \ref{fig:DT}. It consists of two pendulums, pendulum-I modelling the human body part and containing the specimen and pendulum-II representing the robot effective mass and velocity.

\subsubsection{Pendulum-I}

The first pendulum on the left of Fig. \ref{fig:DT} models the human body part by the appropriate effective mass that can be adapted and a specimen mounting allowing for different specimen to be mounted. It is built purely mechanically with low friction bearings. 

Assuming a collision between the robot and a human hand, lower arm, and upper arm simulated based on the effective masses provided by ISO/TS 15066:2016 \cite{iso15066} (\SI{0.6}{kg}, \SI{2}{kg}, \SI{3}{kg}) we attach for experiment condition I-a weights of \SI{1}{kg} arranged in line of motion in order to simulate the effective mass of human hand and lower arm and for I-b a weight of \SI{4}{kg}. These resulting effective pendulum masses are explained following and listed in Table \ref{tab:eff_mass}. 
The pig dew claws or middle feet are mounted at the impact location. Due to the variations in the pig specimen, the final effective mass varies up to \SI{0.3}{kg} extra load. The pendulum's basic structure is fixed with \SI{20}{kg} load to the ground after aligning the collision location to pendulum-II.


\subsubsection{Pendulum-II}

Pendulum-II, depicted on the right of Fig. \ref{fig:DT}, models the robot by its effective mass and allows for various contact geometries to be attached. It is fully automatized and operates control and data acquisition on a compactRio system from the company National Instruments with sampling rate \SI{2000}{Hz}. The mechanical structure consists of a pendulum rod with threaded rods to mount calibrated slit weights of up to \SI{8}{kg}. 
On the contact point, a F/T sensor K6D40 (from the company ME Messsysteme) \cite{force_datasheet} is installed which samples the force sensor data at \SI{300}{Hz} using the corresponding measurement counter GSV-8DS \cite{gsv_datasheet}. On top of the F/T sensor, different impactors can be mounted. In this study, we chose the ones listed in Table \ref{tab:Experimental parameters} made from aluminium alloy EN AW-7075 with hardness \SI{150}{HB} \footnote{Note that these are exactly the same as applied in \cite{Kirschner_2024_Towards}.}

A rotary encoder of type TMCS-28-1k-KIT from ADI Trinamics \cite{encoder_datasheet} measures the pendulum's rotational velocity. A disc brake actuated by a stepper motor is used to stop the pendulum's motion to a) position the pendulum at the appropriate height and b) stop the pendulum to prevent double hitting after recoiling. Lifting the pendulum is operated via a belt, wind, and motor to reach a defined deflection angle $\alpha$, where the desired collision velocity $v_{d}$ is defined by a calibration curve for $\dot{\alpha}$ defined for this device as follows

\begin{equation}
v_{d} = \dot{\alpha}l_{col,II} \, , 
\end{equation} 
\begin{equation}
\dot{\alpha} =  3.8122 \sqrt{1-cos(1.1628\alpha)} \, .
\end{equation}

The calibration curve for the angular velocity is experimentally obtained using the mean angular velocity measured by the rotary encoder for velocities $\geq$ \SI{0.5}{m/s}. The velocity measurements are verified by a light barrier 203.10 with speed counter MZ373 by the company Hentschel System GmbH \cite{Hentschel_datasheet}.

Both pendulums' effective masses at the respective point of collision are calculated by
\begin{equation}
m_\mathrm{p,eff}=  \frac{J_{xx}^{\mathrm{(S)}}+m_\mathrm{p}l^2}{l_\mathrm{col}^2}  \, ,
\end{equation} 
where $J^{\mathrm{(S)}}_{\mathrm{xx}}$ is the inertia around the pendulum center of gravity, $m_\mathrm{p}$ the pendulum's summed mass, $l $ the distance to the center of gravity, and $l_\mathrm{col}$ the distance to the point of collision \cite{Kirschner_2021_Notion}. 
Consequently, we obtain the effective masses in Table \ref{tab:eff_mass} for pendulum-I and pendulum-II with impactor W.

\begin{table}[]
    \centering
    \caption{Modelled pendulum effective masses}
    \begin{tabular}{ccccccc}
    
    \toprule
        pendulum & load & $J_{xx}^{\mathrm{(S)}}$ & $l$ & $l_{col} $ & $m_\mathrm{p}$ & $m_\mathrm{p,eff} $  \\
         & $[kg]$ & $[kgmm^2]$ & $[mm]$ & $[mm]$ & $[kg]$ & $[kg]$ \\
         \midrule
        I& 1 &  409,604.47 & 518 & 794 & 4.54 & 2.58 \\
        I& 4 & 648,362.99 & 670 & 794 & 7.49 & 6.36 \\
        II & 2 & 531,378.16 & 782 & 990 & 4.26 & 3.20 \\
        II & 4 & 583,728.08 & 842 & 990 & 6.26 & 5.12 \\
        II & 6 & 612,297.60 & 873 & 990 & 8.26 & 7.04 \\
        II & 8 & 631,599.33 &  892 & 990 & 10.26 &  8.97\\
         \bottomrule
    \end{tabular}
    
    \label{tab:eff_mass}
    \vspace{-5mm}
\end{table}

\subsubsection{Experimental Protocol}

The experimental protocol is divided in three steps as follows.

\textbf{Preparation procedure:} We thaw the specimens 12 hours before testing at room temperature ($\sim$ $23^\circ C$). Then, we dissect the dew claws from the middle foot (c.f. Fig. \ref{fig:specimen}). The specimen are stored moistened to prevent the skin from drying. The weights on the pendulums are adjusted to the desired loads, the desired impactor is attached to pendulum-II, and finally the specimen to be tested is attached to pendulum-I using a cord tying system.

\textbf{Experimental Phase:}
The experimental impact procedure is depicted in Fig. \ref{fig:protocol_flowchart} and all steps are briefly listed as follows.
First, pendulum-I and -II are manually aligned such that the specimen is hit perpendicularly at the desired collision point, c.f. Fig. \ref{fig:protocol_flowchart} i). We then set the desired collision speed and length of the pendulum to the collision point, $l_{col}$, and start an automated single test run using the developed LabView program. Following, the pendulum is pulled up to the required deflection angle, $\alpha$, by motor, wind, and strap, c.f. Fig. \ref{fig:protocol_flowchart} ii). The pendulum brake is triggered when $\alpha$ is reached. Then, the motor unwinds the strap to allow free falling, c.f. Fig. \ref{fig:protocol_flowchart} iii). Once the strap is unwound, encoder and force sensor measurement is started and the brake is released. The pendulum falls and accelerates to the desired velocity in the collision point where pendulum-II hits the specimen tied to pendulum-I, c.f. Fig. \ref{fig:protocol_flowchart} iv). Both pendulums deflect as a result of the collision. The swing of pendulum-I is caught manually, while pendulum-II uses the automated brake to catch its back swing, c.f. Fig. \ref{fig:protocol_flowchart} v). 

\textbf{Postprocessing:}
After each experiment, the specimen is untied, the collision area marked with red skin marker and a macroscopic optical and tactile evaluation for tissue injuries, such as skin imprints (s-i) or skin cuts (s-c) is conducted, and the tissue structure is documented photographically.  
If clear cuts or palpable structural changes are detected, the injury location is further investigated for tendon/bone (t/b) injuries. Then, the next specimen or collision location is chosen and the specimen tied to pendulum-I. 



\section{Results} \label{sec:results} 

In this section, we present the results for all five descriptive questions listed in Sec. \ref{sec:methodology} and provide brief interpretations. 
In the experiments, the velocity at impact was always measured. With respect to the desired velocity, we obtained the accuracy of the collision velocity based on the desired velocity as: 
\begin{itemize}
    \item $v_d =$\SI{0.25}{m/s}: $v_{p,II} =$\SI{0.34}{m/s} $\pm$ \SI{0.10}{m/s}
    \item $v_d =$\SI{0.5}{m/s}: $v_{p,II} =$\SI{0.54}{m/s} $\pm$ \SI{0.06}{m/s}
    \item $v_d =$\SI{1.0}{m/s}: $v_{p,II} =$\SI{1.03}{m/s} $\pm$ \SI{0.01}{m/s}
    \item $v_d =$\SI{1.5}{m/s}: $v_{p,II} =$\SI{1.56}{m/s} $\pm$ \SI{0.03}{m/s}
    \item $v_d =$\SI{2.0}{m/s}: $v_{p,II} =$\SI{2.03}{m/s} $\pm$ \SI{0.03}{m/s}
\end{itemize}

\begin{figure}[t]
\centering
		\includegraphics[width=1\linewidth]{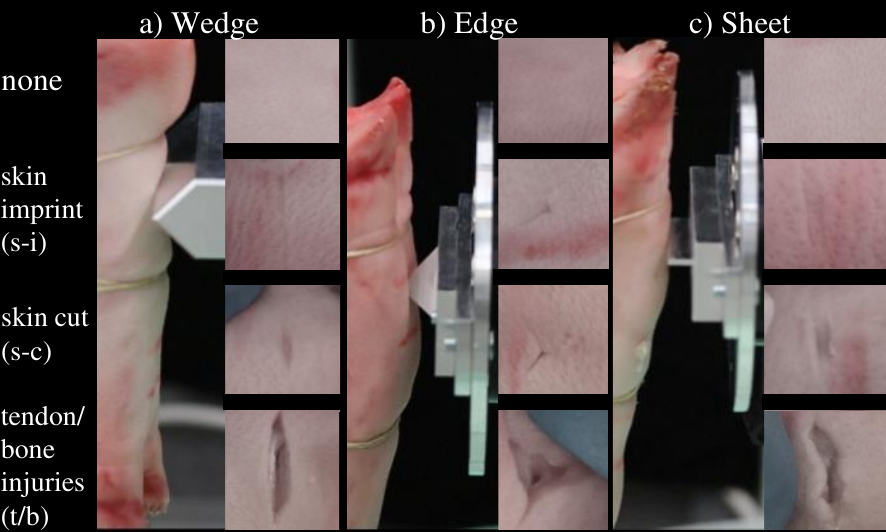}
		
	\caption{Observed features representing injury types occurring upon impacting the pig skin on the dew claws and middle feet with the impactors defined in Table \ref{tab:Experimental parameters}.}
	\label{fig:pig_injuries}
	\vspace{-5mm}
\end{figure}

\subsection{Injury types}

Within all 720 experiments, we observed and grouped the injury types depicted exemplarily in Fig. \ref{fig:pig_injuries}. In total, 6 experiments had to be declared invalid due to previous skin abnormalities in the impacted region. In experimental setting I-a) (360 single experiments), a total of 38 scenarios occurred where no injury was observed, in 156 cases, s-i occurred, 161 times a s-c was observed, and in 5 impacts, a resulting deeper tissue injury such as t/b injuries were observed. In experimental setting I-b) (360 single experiments), a total of 56 scenarios occurred where no injury was observed, in 112 cases, s-i occurred, 172 times a s-c was observed, and in 14 impacts, a resulting deeper tissue injury such as t/b injuries were observed. When t/b was denoted, the impactor had pierced through the skin and caused imprints or even cuts in the periosteum and imprints in the bone itself, which can be classified as compression fractures. Unlike the previous study in \cite{Kirschner_2024_Towards}, no full or fragmented bone fractures were caused under the tested impact conditions. 

\subsection{Injury probability}
\begin{figure}[th]
	
		\includegraphics[width=1\linewidth]{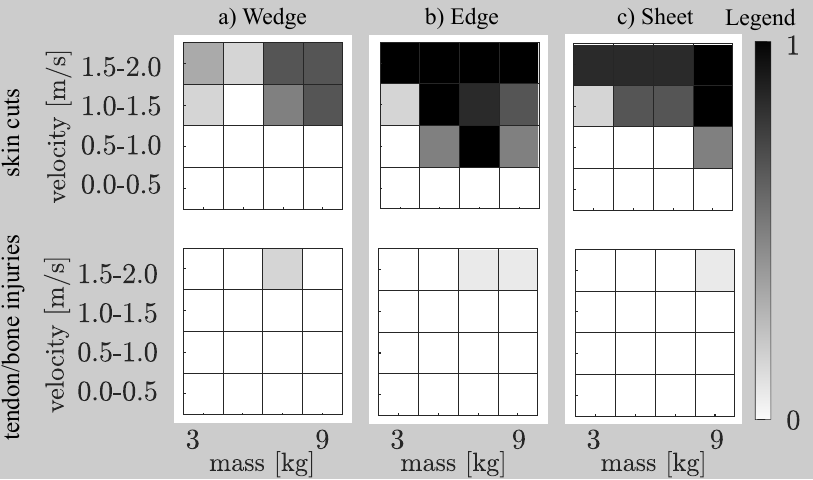}
	\caption{Results for experiment condition I-a, \SI{2.6}{kg} effective mass of pendulum-I (resembling the human): Probability of skin cuts and tendon/bone injuries in unconstrained impacts.}
	
	\vspace{-5mm}
	\label{fig:low_mass_risk_results}
\end{figure}

\begin{figure}[th]
	
		\includegraphics[width=1\linewidth]{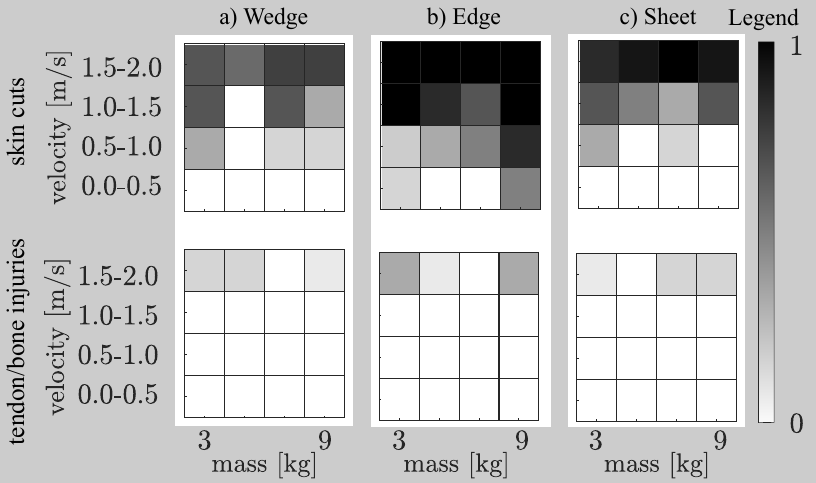}
	\caption{Results for experiment condition I-b, \SI{6.4}{kg} effective mass of pendulum-I (resembling the human): Probability of skin cuts and tendon/bone injuries for unconstrained impacts.}
	
	\vspace{-3mm}
	\label{fig:high_mass_risk_results}
\end{figure}

Based on our experiments, we calculate the probability of occurrence of different injury types as depicted in Fig. \ref{fig:high_mass_risk_results} and \ref{fig:low_mass_risk_results}. These injury probability maps provide an overview of the probability of certain injury types occurring with respect to the effective masses of pendulum-II rounded to the next higher value and the collision velocities, grouped for a range of \SI{0.5}{m/s} each. Table \ref{tab:risk_results} additionally lists the probability of s-c or t/b injuries occurring considering robots with effective masses between \SI{3}{}-\SI{9}{kg} which are limited to a) \SI{2.0}{m/s}, b) \SI{1.0}{m/s}, and c) \SI{0.5}{m/s} Cartesian motion speed for all three impactors and two experimental series. In Sec. \ref{sec:Discussion}, we discuss the application of the injury probability knowledge for pHRI risk assessments.

\begin{table}[]
    \centering
    \caption{Overall injury probability for tested impactors for entire parameter space $m_r =$ \SI{3}{}-\SI{9}{kg} up to \SI{2}{m/s} vs. reduced velocity up to \SI{1}{m/s} and \SI{0.5}{m/s}}
    \begin{tabular}{p{5mm}p{9mm}p{5mm}P{8mm}P{5mm}P{8mm}P{5mm}P{5mm}}
    
    \toprule
    & & \multicolumn{2}{c}{ $v_r <$ \SI{2}{m/s}} & \multicolumn{2}{c}{$v_r <$ \SI{1}{m/s}} & \multicolumn{2}{c}{$v_r <$ \SI{0.5}{m/s}}\\
        $m_{pI}$ & impactor & s-c [\%] & t/b [\%] & s-c [\%] & t/b [\%] & s-c [\%] & t/b [\%] \\
         \midrule
         
        \multirow{2}{*}{I-a} & W &  19.8  & 1.0 & 0.0 & 0.0 & 0.0 & 0.0 \\
         & E & 54.2 & 1.0  & 25.0 & 0.0 & 0.0 & 0.0  \\
         & S & 40.6 & 0.5  & 6.3 & 0.0 & 0.0 & 0.0 \\
        \multirow{3}{*}{I-b} & W & 31.8 & 2.6  & 8.3 & 0.0 & 0.0 & 0.0 \\
         & E & 62.7 & 4.7  & 31.7 & 0.0 & 16.7 & 0.0 \\
         & S & 39.6 & 2.6  & 6.3 & 0.0 & 0.0 & 0.0 \\
         \bottomrule
    \end{tabular}
    \vspace{-5mm}
    \label{tab:risk_results}
\end{table}

\subsection{Injury forces}
\begin{figure}[th]
	
		\includegraphics[width=1\linewidth]{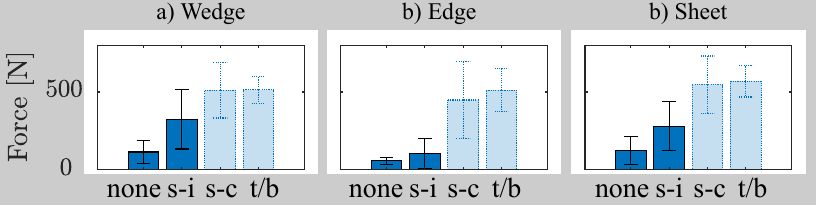}
	\caption{Forces measured during experiment series I-b.}
	
	\vspace{-5mm}
	\label{fig:forces_results}
\end{figure}

The measured forces that cause different types of injury are shown for the test arrangement I-b \footnote{values are not shown due to a sensor error in the experiments with I-a} in Fig. \ref{fig:forces_results} is shown. Between the impactors, the forces that did not cause macroscopically visible injuries were \SI{114}{} $\pm$ \SI{75}{N} for impactor W, \SI{57}{} $\pm$ \SI{27}{N} for impactor E, \SI{122}{} $\pm$ \SI{89}{N} for impactor S. S-i was visible at \SI{325}{} $\pm$ \SI{179}{N} for impactor W, \SI{105}{} $\pm$ \SI{97}{N} for impactor E, \SI{280}{} $\pm$ \SI{158}{N} for impactor S. Skin cuts occurred at \SI{514}{} $\pm$ \SI{89}{N} for impactor W, \SI{448}{} $\pm$ \SI{246}{N} for impactor E, \SI{568}{} $\pm$ \SI{100}{N} for impactor S. The values above \SI{500}{N} should be treated with caution, as the official measuring range of the sensor is exceeded.
While there are clear differences between the forces that may be associated with no injuries, s-i, or s-c, the distinction between skin cuts and more severe tissue injuries such as t/b injuries is not as clear. This artifact may be due to the anatomical differences between the different finger positions. Only in the case of the distal dew claws did such injuries occur, and here the skin is thinner and more stretched over the phalanx bone, exposing the bone to more force than in the other locations. The injury forces measured in this experiment correspond to initial impact forces with a duration of $<<$ \SI{0.5}{s} and can be expected in human-robot contacts under similar circumstances. Please note that the comparison of injury forces between the study presented here and the restricted injury study in \cite{Kirschner_2024_Towards} is not recommended, as the force sensors used have a different measurement range and sampling frequency.

\subsection{Injury Protection datasets} 
\begin{figure*}[th]
	
		\includegraphics[width=1\linewidth]{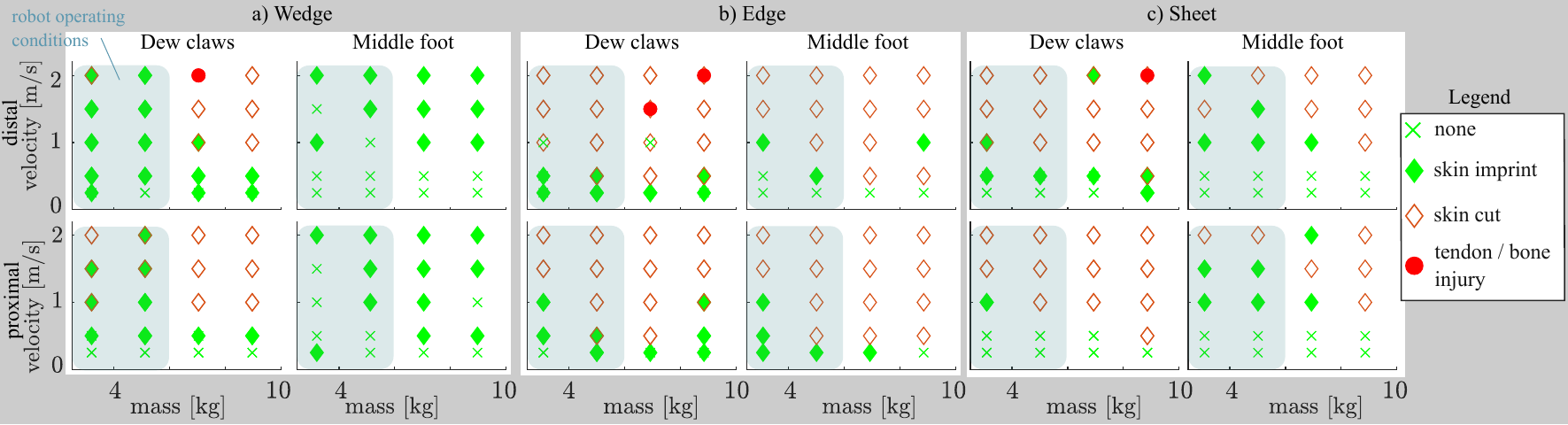}
	\caption{Results for \SI{2.6}{kg} effective pendulum-I mass, Overview of the observed injuries occurring during the experimental unconstrained impacts. As application example, the probable operating conditions for a FE robot are included.}
	
	\vspace{-3mm}
	\label{fig:low_mass_results}
\end{figure*}

\begin{figure*}[th]
	
		\includegraphics[width=1\linewidth]{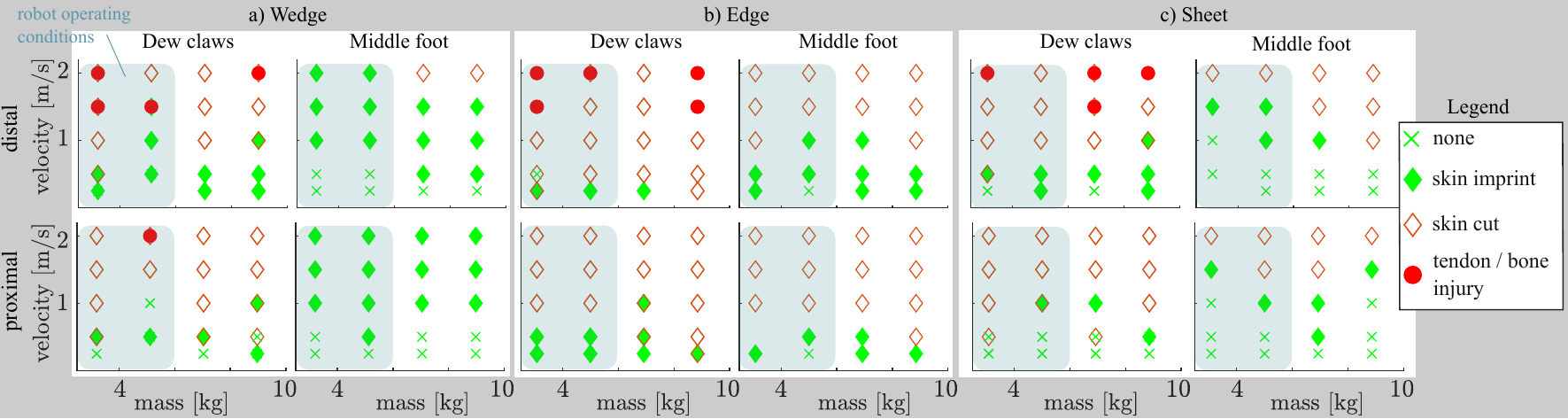}
	\caption{Results for \SI{6.4}{kg} effective pendulum-I mass, Overview of the observed injuries occurring during the experimental unconstrained impacts. As application example, the probable operating conditions for a FE robot are included. }
	
	\vspace{-6mm}
	\label{fig:high_mass_results}
\end{figure*}

In Fig. \ref{fig:low_mass_results} and Fig. \ref{fig:high_mass_results}, we compile injury protection datasets for both effective human mass assumptions considered in this study. These datasets can be used to perform risk assessments for human-robot interaction and perform risk mitigation by speed reduction according to the injury severity found in the database in combination with the computable reflected inertia of the robot similar to \cite{Haddadin.2012}. As an example, typical operating conditions for a FE robot with a reflected inertia between $\sim$ \SI{3}{}-\SI{6}{kg}, determined in \cite{Kirschner_2021_Notion}, and velocities up to \SI{2}{m/s} are given.  

As expected, impactor E caused the most severe injuries in our study due to the very small collision area and shows the most s-c and t/b injuries. Nevertheless, at an impact velocity of \SI{0.25}{m/s}, no injuries or s-i were visible on the pig dew claws and mid feet in most cases. At W and S, we observed less severe injuries at the back of the pig's feet than at the dew claws. This probably results from the larger collision area and the associated lower energy density in the individual points on impact with the large area of the back of the pig's feet. The pig's metatarsae have a similar structure to the dorsal side of the distal human forearms and hands, while the dew claws are more like human fingers. Differentiating between a collision with a finger or the hand prior to a collision scenario therefore theoretically allows the robot's movement speed to be increased by up to 8 times (if skin impressions are considered an acceptable injury).

\subsection{Comparison of experiment I-a, I-b, and constrained tests}
When comparing experimental settings I-a, \SI{2.6}{kg} effective mass of pendulum I, and I-b, \SI{6.4}{kg}, we observed only few differences. The probability of skin cuts increased by more than 10\% for impactors W and E, and the probability of t/b injury increased by more than 1\%. For impactor S, only the probability of t/b injury increased, by 2\%. Given the qualitative variances in skin structure between different pig specimens and the expected differences to human tissue, these changes appear to be of low significance. Therefore, due to the exploratory nature of this study, we conclude that a difference in human effective mass of approximately \SI{3}{kg} likely does not cause significant differences in the severity of injuries. However, further investigation with more specimens under the same robot mass-velocity conditions is required.


The contact between a human in constrained condition, scenario c), can be considered as a scenario a) contact where the human effective mass becomes infinite ($m_{h} \approx \infty$) \cite{Svarny_2021,Haddadin.2009_requ}. To gain an impression of the effect the human body part mass causes, we thus compare to our previous work in \cite{Kirschner_2024_Towards}, where we investigated pig dew claws in constrained collisions using the same impactors. While the tested scenario in this work which uses a drop test device is scenario a) and b) combined, we can assume due to the low forces resting on the specimen (max. \SI{30}{N}) that injury was not caused due to scenario b). In \cite{Kirschner_2024_Towards}, the impact masses causing only s-i even at low velocities were mostly below \SI{4}{kg} and velocities $\geq$ \SI{0.25}{m/s} easily caused skin cuts. In the study on unconstrained collision in this paper, we see clearly that the robot masses and velocities that cause injury in unconstrained collisions are more than double as high for all impactors\footnote{Note that, due to differences in force sensing sampling rate in comparison to \cite{Kirschner_2024_Towards}, we refrain from comparing the results on force causing injuries.}. 


\section{Discussion} \label{sec:Discussion}

This section outlines potential applications of the generated data, discusses the contributions and limitations of this initial study on injuries in unconstrained collisions, and suggests future work.

In the electronics industry, collaborative assembly involves handling light-weight, edged printed circuit boards (PCBs). ISO 12100:2011 requires a risk assessment based on machinery limitations. For example, using a Franka Emika (FE) robot for tactile fitting, we can calculate the effective mass (typically 3-6 kg) and a maximum Cartesian motion speed of \SI{2}{m/s}. To mitigate risk, robot motion space constraints and physical barriers can prevent constrained contact in undesired locations. Injury severity and probabilities can be estimated using Table \ref{tab:risk_results}. If needed, mass-velocity information in Fig. \ref{fig:low_mass_results} or \ref{fig:high_mass_results} can limit injury potential by adjusting robot configurations and motion speed. Alternatively, real-time capable control can ensure safe and efficient robot motion.

This study outlines the necessary experimental setups and procedures to gather data that complement existing human injury research. These exploratory experiments serve as a foundation for future studies but do not support statistical hypothesis testing. Future statistical testing should include a better understanding of effective body part masses, extending data from ISO/TS 15066:2016 with kinematic and dynamic models and \emph{in vivo} studies. Additionally, while this study focuses on normal contacts to pig skin, future research will consider shearing forces from angular contact.

With perspective on our experimental setup, some limitations were found during this study. For velocities $\geq$ \SI{0.25}{m/s}, the pendulum deflection may only be minimal. The currently implemented direct drive of the wind does not allow these small angles with high precision. Future versions of the test stand will include a gear coupling to improve the low velocity precision. Furthermore, manual positioning and aligning of pendulum-I cause offsets in the collision velocity, which need to be reduced in future test rigs by automated positioning. The unambiguous comparison of the collision forces causing injury between the previous study on constrained collision and this one on unconstrained contacts requires comparable force sensor performance, i.e. similar measuring rate and range, which was not the case so far.

More generally, the observed skin imprints may  result in haematoma (contusions), which our study cannot unambiguously interpret. Investigation of contusion requires additional \emph{in vivo} studies similar to \cite{Behrens2023}. Pig feet seem promising as human surrogates due to their anatomical similarity and availability for large-scaled studies. Nevertheless, to understand the limitations of their applicability as human hand and arm surrogates requires comparative human studies. Based on such human arm studies, a general classification scheme for less severe skin, soft tissue, and superficial bone injuries is required extending existing medical classifications to ensure a comparable classification in further studies.


\section{Conclusion} \label{sec:Conclusion}
\vspace{-0.1cm}
This work extends previous investigations on injury potential in pHRI by analyzing injuries from unconstrained collisions. We built an experimental setup with two pendulums to model the effective masses of the human body parts and robots. The actuated pendulum, representing the robot, can be equipped with different impact geometries and achieve specific velocities, while the passive pendulum holds \emph{ex vivo} specimens (pig feet) as human surrogates. We conducted 720 experiments using various effective masses (\SI{3}{}-\SI{9}{kg}) and velocities (\SI{0.25}{}-\SI{2.0}{m/s}) to explore injury potential in light, unconstrained collisions with edged or pointed geometries. Results show significant differences from previous constrained collision tests, highlighting the need for differentiated collision scenarios in planning or real-time control of safe human-robot interaction. The higher injury potential observed on finger substitutes and the back of the hand suggests further work on distinct body part recognition. Lastly, to ensure safe robot deployment near humans, we need to understand soft tissue injuries from human-robot collisions. By enhancing datasets and conducting comparative \emph{ex vivo} studies, we aim to build this knowledge. 

\section*{ACKNOWLEDGMENT}
\vspace{-0mm}
We thank Fritz Seidl for his support and gratefully acknowledge the funding of the Lighthouse Initiative KI.FABRIK Bayern by StMWi Bayern, Forschungs- und Entwicklungsprojekt, grant no. DIK0249 and of the Bavarian State Ministry for Economic Affairs, Regional Development and Energy (StMWi) as part of the project SafeRoBAY (grant number: DIK0203/01).
 \vspace{-1mm}

\bibliographystyle{IEEEtran} 
\bibliography{References/literature.bib}

\end{document}

%% file: Settings/settings.tex
\IEEEoverridecommandlockouts                              

\usepackage{footnote}
\usepackage{amsmath} 
\usepackage{amssymb}  
\usepackage{booktabs}
\usepackage{color}
\usepackage{xcolor}
\usepackage{multirow}

\usepackage[pdftex]{graphicx}
\usepackage{siunitx}
\usepackage{cite}

\usepackage{hyperref}
\usepackage{subcaption}
\usepackage{glossaries}
\include{Settings/glossary}

\definecolor{excelblue}{HTML}{4472C4} 
\definecolor{NewBlue}{HTML}{4472C4} 
\definecolor{NewOrange}{HTML}{ED7D31} 
\definecolor{NewGreen}{HTML}{00B050} 

\usepackage{stackengine}
\newcolumntype{P}[1]{>{\centering\arraybackslash}p{#1}}

\usepackage{fancyhdr}
\input{Settings/abkuerzung.sty}
\graphicspath{{graphics/}}
\DeclareGraphicsExtensions{.pdf,.jpeg,.png}

\renewcommand{\baselinestretch}{0.94}

\def\mytitle{Towards Unconstrained Collision Injury Protection Data Sets: Initial Surrogate Experiments for the Human Hand}
\title{\LARGE \bf \mytitle}

\def\myauthor{Robin Jeanne Kirschner$^{1}$, Jinyu Yang$^{1}$, Edonis Elshani$^{1}$, Carina M. Micheler$^{2,3}$,  Tobias Leibbrand$^{1}$, \\Dirk Müller$^{2}$, Claudio Glowalla$^{2,4}$, Nader Rajaei$^{1}$, Rainer Burgkart$^{2}$ and Sami Haddadin$^{1}$}
\def\mythanks{$^1$ Chair for Robotics and Systems Intelligence, Munich Institute of Robotics and Machine Intelligence, Technical University of Munich, 80992 Munich, Germany\newline 
$^2$ Department of Orthopaedics and Sports Orthopaedics, Klinikum rechts der Isar, TUM School of Medicine, Technical University of Munich, 81675 Munich, Germany\newline
$^3$ Institute for Machine Tools and Industrial Management, TUM School of Engineering and Design, Technical University of Munich, 85748 Garching near Munich, Germany\newline
$^4$  Endoprothetikzentrum der Maximalversorgung, BG Hospital Murnau, 82418 Murnau, Germany\newline
Corresponding author:
{\href{mailto:robin-jeanne.kirschner@tum.de}{\tt\small robin-jeanne.kirschner@tum.de}}
}

\author{\myauthor
\thanks{}
\thanks{\mythanks}%
}

\hypersetup{ 
pdftitle={\mytitle},
pdfauthor={\myauthor},
}

%% file: Settings/glossary.tex
\newacronym{HRI}{HRI}{human-robot interaction}
\newacronym{EBC}{EBC}{Expectation-Based Control}
\newacronym{MPC}{MPC}{Model Predictive Control}
\newacronym[longplural={degrees of freedom},shortplural={DoFs}]{DoF}{DoF}{degree of freedom}
\newacronym{SMU}{SMU}{Safe Motion Unit}
\newacronym{SSM}{SSM}{speed and separation monitoring}
\newacronym{PFL}{PFL}{power and force limiting}
\newacronym[longplural={inertial measurement units},shortplural={IMUs}]{IMU}{IMU}{inertial measurement unit}
\newacronym[longplural={evasive motions},shortplural={EMs}]{EM}{EM}{evasive motion}
\newacronym[longplural={Expectable Motion Units},shortplural={EMUs}]{EMU}{EMU}{Expectable Motion Unit}
\newacronym{FE}{FE}{Franka Emika}
\newacronym{IMO}{IMO}{human involuntary motion occurrence}
\newacronym{TUI}{TUI}{Technology Usage Inventory}
\newacronym{PS}{PS}{perceived safety}
\newacronym{GS}{GS}{Godspeed}